\documentclass[runningheads]{llncs}
\usepackage[T1]{fontenc}
\usepackage[hidelinks]{hyperref}
\usepackage{booktabs}
\usepackage{algorithm}
\usepackage{algorithmic}
\usepackage{mathtools}
\usepackage{enumitem}
\usepackage{amssymb,dsfont}
\usepackage{tabularx}
\usepackage{multirow}

\usepackage{caption}
\usepackage{subcaption}
\usepackage{comment}
\usepackage{graphicx}
\usepackage{csquotes}
\usepackage{enumitem}
\usepackage{nth}
\usepackage{makecell}
\usepackage{bm} 

\usepackage{tikz}
\usetikzlibrary{positioning, shapes.geometric}

\usepackage{cleveref}

\usepackage{placeins}

\usepackage{xcolor}
\definecolor{plotBlack}{RGB}{0,0,0}
\definecolor{plotRed}{RGB}{255,0,0}
\definecolor{plotBlue}{RGB}{0,0,255}
\definecolor{plotGreen}{RGB}{0,128,0}

\renewcommand{\paragraph}[1]{\par\noindent\textbf{#1}\,}

\makeatletter
\usepackage{tikz}

\makeatother

\usepackage{graphicx}
\usepackage{color}

\title{Uncertainty Gating for Cost-Aware Explainable Artificial Intelligence}

\begin{document}
%
%
\titlerunning{Uncertainty Gating for Cost-Aware Explainable Artificial Intelligence}

\author{
Georgii Mikriukov\inst{1}\orcidID{0000-0002-2494-6285}
\and Grégoire Montavon\inst{2,3}\orcidID{0000-0001-7243-6186}
\and Marina M.-C. Höhne\inst{1,4}\orcidID{0000-0003-3090-6279}
}

\authorrunning{G. Mikriukov et al.}

\institute{
Leibniz Institute for Agricultural Engineering and Bioeconomy (ATB), Potsdam, Germany\\
\email{\{GMikriukov,MHoehne\}@atb-potsdam.de}
\and
BIFOLD -- Berlin Institute for the Foundations of Learning and Data, Berlin, Germany
\and
Charité -- Universitätsmedizin Berlin, Berlin, Germany\\
\email{gregoire.montavon@charite.de}
\and
University of Potsdam -- Department of Computational Science, Potsdam, Germany
}

%
\maketitle              
\setcounter{footnote}{0}  

\sloppy

\begin{abstract}
Post-hoc explanation methods are widely used to interpret black-box predictions, but their generation is often computationally expensive and their reliability is not guaranteed. We propose epistemic uncertainty as a low-cost proxy for explanation reliability: high epistemic uncertainty identifies regions where the decision boundary is poorly defined and where explanations become unstable and unfaithful. This insight enables two complementary use cases: `improving worst-case explanations' (routing samples to cheap or expensive XAI methods based on expected explanation reliability), and `recalling high-quality explanations' (deferring explanation generation for uncertain samples under constrained budget). Across four tabular datasets, five diverse architectures, and four XAI methods, we observe a strong negative correlation between epistemic uncertainty and explanation stability. Further analysis shows that epistemic uncertainty distinguishes not only stable from unstable explanations, but also faithful from unfaithful ones. Experiments on image classification confirm that our findings generalize beyond tabular data.$^{\star}$


\end{abstract}

\begingroup
\renewcommand{\thefootnote}{\fnsymbol{footnote}}
\footnotetext[1]{The code is available at \url{https://github.com/comrados/uq-xai}.}
\endgroup

\section{Introduction}
\label{sec:introduction}

Explainable AI (XAI) methods have become essential for interpreting black-box model predictions in high-stakes domains~\cite{adadi2018peeking,arrieta2020explainable}. In particular, `model-agnostic' explanations~\cite{lundberg2017unified,ribeiro2016should} have proven flexible across a wide range of models, but this comes with substantial computational cost and unreliable outputs: explanations may inherit uncertainty from the model or arise from biases intrinsic to the explanation technique, often in an input-dependent manner. Existing evaluation criteria such as faithfulness and stability~\cite{yeh2019fidelity,adebayo2018sanity} can detect unreliable explanations, but only retrospectively and at high computational cost, making them impractical at runtime. Thus, current XAI lacks a low-cost mechanism to decide how much computational effort to invest in an explanation before generating it.
 
Uncertainty quantification (UQ) addresses model reliability by identifying regions where predictions are poorly supported by training data. Recent works connect UQ with XAI -- using explanations to estimate uncertainty~\cite{salvi2025explainability}, analyzing how uncertainty undermines explanation stability~\cite{chiaburu2025uncertainty}, or explaining uncertainty itself~\cite{bley2025explaining} -- yet all still require generating explanations first. Most closely related, \cite{zhu2025robust} uses uncertainty to guide explanation type selection or reject unreliable explanations. However, no prior work frames uncertainty-guided XAI resource allocation as a cost-benefit problem, where deferring explanations yields computational savings or prevents misleading attributions.
 
In this paper, we systematically explore the relationship between epistemic uncertainty and explanation reliability across multiple datasets, architectures, and XAI methods. We hypothesize that explanations are inherently unstable and unfaithful in high epistemic uncertainty regions, enabling two complementary use cases: (1)~\emph{improving worst-case explanations}, where epistemic uncertainty routes samples to low- or high-cost XAI methods depending on expected reliability, and (2)~\emph{recalling high-quality explanations}, where explanation generation is deferred for high-uncertainty samples under constrained computational budget.
For models without low-cost native epistemic uncertainty estimation, lightweight surrogate models provide effective proxies.
Our contributions are as follows:
\begin{itemize}[nosep,label=\textbullet]
    \item \emph{UQ-XAI correlation analysis:} We identify systematic relationships between epistemic uncertainty and XAI stability and faithfulness, including a clear monotonic pattern across epistemic strata (low $>$ medium $\gg$ high).
    \item \emph{Uncertainty-guided resource allocation:} We propose a framework that routes explanation effort based on epistemic uncertainty and quantify the resulting cost-quality trade-offs across XAI methods and deferral regimes.
    \item \emph{Cost-efficient epistemic proxies:} We demonstrate that lightweight uncertainty estimators (e.g., random forest) provide sufficiently strong signals to enable reliable gating at negligible overhead relative to XAI cost.
\end{itemize}

\section{Related Work}
\label{sec:related}

Our work lies at the intersection of uncertainty quantification, XAI, and AI robustness under perturbations, with a focus on tabular data. The related work can be organized in three categories.

\paragraph{Methods for Uncertainty Quantification} 
Uncertainty quantification (UQ) distinguishes \emph{aleatoric uncertainty}, induced by inherent data noise, from \emph{epistemic uncertainty}, reflecting limited model knowledge in underrepresented regions~\cite{hullermeier2021aleatoric,kendall2017uncertainties}. For ensemble-based models, epistemic uncertainty is typically computed as prediction variance across ensemble members:

\begin{align}
\mathcal{U}_{\text{epi}}(\mathbf{x}) = \text{Var}_{m}[f_m(y=k|\mathbf{x})],
\label{eq:ensemble-epistemic}
\end{align}

Established methods include prediction variance for random forests~\cite{shaker2020aleatoric}, deep ensembles and MC Dropout for neural networks~\cite{lakshminarayanan2017simple,gal2016dropout}, surrogate-based approaches for gradient boosting~\cite{malinin2021uncertainty,duan2020ngboost}, and bootstrap resampling for linear models~\cite{efron1994introduction}.

In contrast to these works, we do not propose a new method for estimating predictive uncertainty but use the latter as a criterion for deciding how much effort is required to explain those predictions.

\paragraph{Evaluating Explanation Methods}
As XAI typically cannot be simply assessed based on ground-truth explanations, and must also fulfill complex multifaceted desiderata \cite{swartout1993}, significant work has been dedicated to the question of how to assess explanation quality and usefulness, summarized in reviews such as \cite{nauta2023}. As major families of evaluation techniques, we can mention the `pixel-flipping' and insertion/deletion metrics \cite{samek2017evaluating}, as well as measures of explanation stability \cite{montavon2018,alvarez2018robustness,mikriukov2023evaluating}. Multiple works document explanation vulnerability to perturbations and adversarial attacks~\cite{baniecki2024adversarial,mikriukov2024unveiling}.

In comparison to these classical methods for evaluating explanations, which often require repeatedly calling the ML model with XAI-derived input perturbations, we establish a fast surrogate that is purely based on predictive uncertainty and that can be computed quickly for single data points.

\paragraph{Connections between UQ and XAI}
Recent work integrates UQ with XAI. Bayesian extensions of LIME and SHAP model explanation uncertainty internally by producing credible intervals for individual attributions~\cite{zhao2021baylime,slack2021reliable}; a broader review of UQ modeling and human perception of uncertainty in explanations is provided in~\cite{chiaburu2024uncertainty}. Uncertainty decomposition can guide both explanation rejection and explanation type selection~\cite{zhu2025robust}. Uncertainty-aware explanations reduce interpretation biases in healthcare~\cite{salvi2025explainability} and identify uncertainty drivers in manufacturing~\cite{mehdiyev2025quantifying}. Work on uncertainty propagation examines how perturbations in the data and model affect explanation stability~\cite{chiaburu2025uncertainty}, while~\cite{bley2025explaining} proposes a method that explains the prediction uncertainty itself, by identifying input features that contribute to it.

Compared to these works, we provide a different perspective, viewing UQ as a proxy for explanation reliability, enabling both computational gains and quality-aware routing in dataset-wide XAI. Specifically, this is achieved by leveraging strong correlations between epistemic uncertainty and explanation stability and faithfulness.

\section{Epistemic Gating for Cost-Aware XAI}
\label{sec:methodology}

In this section, we propose using epistemic uncertainty as a routing signal to allocate explanation effort based on predicted reliability.

\subsection{The Epistemic Gating Mechanism}

Post-hoc explanation methods vary in cost: KernelSHAP~\cite{lundberg2017unified} requires thousands of model evaluations per sample, while TreeSHAP~\cite{lundberg2020trees} operates in polynomial time. Crucially, not all predictions require equally thorough analysis: samples in well-constrained regions yield reliable explanations with lightweight methods, whereas samples near poorly-defined decision boundaries warrant more thorough attribution. Our framework exploits this asymmetry by routing samples toward the appropriate explanation effort based on epistemic uncertainty.

We consider supervised classification settings where a model 
$f: \mathbb{R}^d \to \Delta^K$ maps an input $\mathbf{x} \in \mathbb{R}^d$ 
to a probability distribution over $K$ classes. Given a test sample, 
we aim to assess whether the post-hoc explanation is expected to be 
faithful and stable under input perturbations, and whether these 
properties are predictable from the epistemic uncertainty of the prediction.

\begin{figure}[!h]
    \centering
    \includegraphics[width=\linewidth]{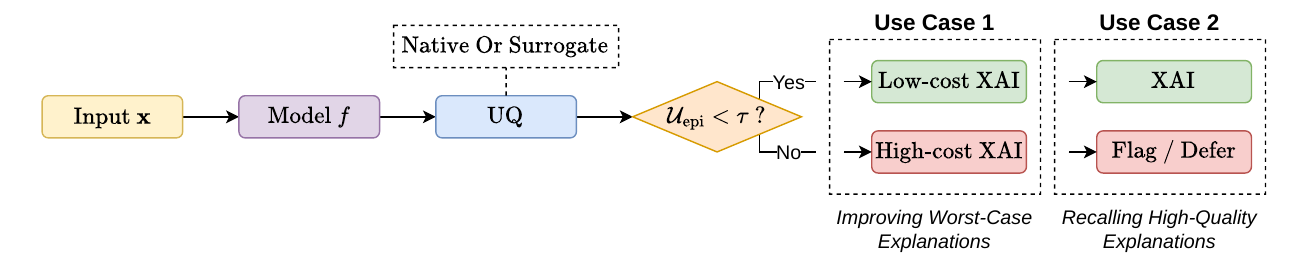}
    \caption{Uncertainty-guided XAI resource allocation via epistemic gating. Epistemic uncertainty is obtained from the model's native estimator or a lightweight surrogate. Under constrained computation budget, \emph{Use Case 1} routes samples to low- or high-cost XAI methods based on expected reliability; \emph{Use Case 2} defers high-uncertainty samples, retaining only reliable explanations.}
    \label{fig:gating}
\end{figure}

Epistemic uncertainty captures model uncertainty from limited data and insufficiently constrained parameters. We hypothesize that high epistemic uncertainty indicates regions where the decision boundary is poorly defined and explanations are prone to instability, making it a natural proxy for local model reliability and a gate for explanation resource allocation. As illustrated in Figure~\ref{fig:gating} (left), for each input $\mathbf{x}$ of model $f$, epistemic uncertainty $\mathcal{U}_{\text{epi}}(\mathbf{x})$ is compared to the threshold $\tau$ to route the sample to the appropriate downstream action. For models lacking low-cost native uncertainty estimation, a lightweight Random Forest surrogate provides a stable proxy signal~\cite{shaker2020aleatoric}.

\subsection{Dimensionality Analysis of Computational Gains}
\label{sec:cost}

A key motivation for epistemic gating is the substantial asymmetry between uncertainty estimation and post-hoc explanation generation. Although epistemic uncertainty can typically be obtained through lightweight ensembling or stochastic inference, model-agnostic XAI techniques require orders of magnitude more model evaluations per sample, making them impractical to apply exhaustively on large datasets. To formalize this asymmetry, let $m$ represent the number of model evaluations required to produce an uncertainty estimate, and $d$ the number required to generate an explanation. If a fraction $\nu$ of samples is deferred by epistemic gating, the total computational cost relative to the baseline is:
\begin{align}
q &= \frac{m}{d} + (1-\nu)
\label{eq:dim1}
\end{align}
If the explained model is itself an ensemble (e.g., a random forest), uncertainty estimation reduces to a byproduct of inference, improving the ratio to:
\begin{align}
q &= \frac{1}{d} + (1-\nu)
\label{eq:dim2}
\end{align}
Cost improvements are drastic when starting from an expensive XAI method (large $d$) and allowing a high deferral rate (large $\nu$). For example, LIME~\cite{ribeiro2016should} requires $d \sim 10^3$--$10^4$ perturbations per sample, while MC Dropout~\cite{gal2016dropout} requires only $m \sim 10^1$--$10^2$ forward passes, yielding $m/d \sim 0.01$. 

More broadly, \emph{epistemic gating ensures that high-cost explanation is incurred only where the model is uncertain}, rather than applied uniformly to all inputs.

\subsection{Exemplary Use Cases}

The right-hand side of Figure~\ref{fig:gating} illustrates two complementary deployment scenarios enabled by the gating mechanism.

\textbf{Use Case 1: Improving Worst-Case Explanations.} When explanations are required for all inputs, epistemic uncertainty guides the \emph{choice of XAI method} rather than the decision to explain. If $\mathcal{U}_{\text{epi}}(\mathbf{x}) < \tau$, a low-cost method suffices; if $\mathcal{U}_{\text{epi}}(\mathbf{x}) \geq \tau$, the sample lies near a poorly-defined decision boundary and a more thorough multi-pass method is required. This routing strategy allocates greater explanation effort precisely where reliability is hardest to achieve.

\textbf{Use Case 2: Recalling High-Quality Explanations.} When computational budget is limited, epistemic uncertainty serves as a \emph{hard gate}: low-uncertainty samples proceed to XAI computation, while high-uncertainty samples are deferred, saving computation on predictions where attributions would be fragile and potentially misleading. 

Beyond these cases, epistemic uncertainty can accompany any explanation as a continuous reliability indicator, helping users contextualize attributions~\cite{zhang2020effect,ma2024you}.

\section{Experimental Setup}
\label{sec:setup}

This section describes our experimental framework: the datasets used for evaluation, the ML models and their uncertainty estimation methods, the XAI techniques under comparison, and the stability metrics and perturbation types.

\subsection{Datasets}
\label{sec:datasets}

We evaluate our framework on diverse sets of tabular classification benchmarks from the UCI Machine Learning Repository\footnote{\url{https://archive.ics.uci.edu/}}, summarized in Table~\ref{tab:datasets}. These datasets span binary and multi-class classification settings with varying sample sizes and feature dimensionalities, and exhibit different degrees of noise and class imbalance, providing a representative testbed for assessing the relationship between epistemic uncertainty and explanation reliability. Data are split randomly (not stratified) into training, validation, and test set (70/15/15), and features are standardized using a z-score normalization fitted on the training split. 

\begin{table}[h!]
\centering
\caption{Dataset statistics and train/validation/test splits used in experiments.}
\label{tab:datasets}
\setlength{\tabcolsep}{4pt}
\fontsize{8pt}{10pt}\selectfont
\smallskip
\begin{tabular}{l | l | c c | c c c}
\hline
\textbf{Type} & \textbf{Dataset} & \textbf{Features} & \textbf{Classes} & \textbf{Train} & \textbf{Val} & \textbf{Test} \\
\hline
\hline
\multirow{4}{*}{Tabular} 
 & Wine Quality & 11 & 2 & 4547 & 975 & 975 \\
 & Dry Bean & 16 & 7 & 9527 & 2042 & 2042 \\
 & Rice & 7 & 2 & 2667 & 571 & 572 \\
 & Ecoli & 7 & 8 & 235 & 50 & 51 \\
\hline
Image & PlantVillage & 128$\times$128$\times$3 & 3 & 3936 & 843 & 847 \\
\hline
\end{tabular}
\end{table}

The \textit{Wine Quality} dataset combines red and white wine samples described by physicochemical properties. Quality scores are binarized into \textit{low} ($\leq 5$) and \textit{high} ($\geq 6$) classes. The \textit{Dry Bean} dataset contains features of different bean varieties, 7 well-separated classes. The \textit{Rice} dataset distinguishes between 2 rice varieties based on grain characteristics.
The \textit{Ecoli} dataset comprises protein localization sites with 8 highly imbalanced classes.
To assess cross-domain generality of our proposed framework, we further consider an image classification task using a subset of the \textit{PlantVillage}\footnote{\url{https://www.kaggle.com/datasets/emmarex/plantdisease}} dataset containing \enquote{healthy}, \enquote{bacterial spot}, and \enquote{late blight} tomato leaf images. Images are resized to $128 \times 128 \times 3$ pixels and normalized to $[-1, 1]$.

\subsection{ML Models}
\label{sec:models}

To ensure architectural diversity, we train five models of different architectures: \textit{Logistic Regression (LR)} serves as a linear baseline with L2-regularization ($C=1.0$). Epistemic uncertainty is estimated using a bootstrap ensemble of 20 resampled models.
\textit{Random Forest (RF)} consists of 100 trees with a maximum depth of 15. Epistemic uncertainty is obtained directly via prediction variance across trees.
\textit{Multi-Layer Perceptron (MLP)} employs a fully-connected network with architecture $d \to 128 \to 64 \to K$, ReLU activations, and dropout rate $p=0.3$ after each hidden layer. Models are trained for up to 100 epochs with Adam optimizer (learning rate $10^{-3}$) and early stopping. Epistemic uncertainty is estimated using MC Dropout with 50 stochastic forward passes.
\textit{Gradient Boosting (GBDT):} \textit{LightGBM (LGBM)} and \textit{CatBoost (CB)} with 200 trees, learning rate 0.1, depth 6-8. As GBDTs lack native epistemic estimates, a RF surrogate provides uncertainty signals.
For the PlantVillage dataset, we train a VGG-like CNN model composed of three convolutional blocks followed by a classification head. Each block consists of a $3\times3$ convolution (pad $1$), ReLU activation, $2\times2$ max pooling, and dropout with rate $p=0.3$. The number of channels increases across blocks from 32 $\to$ 64 $\to$ 128. The feature extractor is followed by global average pooling and a final linear layer producing class logits. 

\subsection{XAI Methods}
\label{sec:implementation}

We benchmark five XAI methods: SHAP (TreeExplainer / KernelExplainer)~\cite{lundberg2017unified,lundberg2020trees}, LIME~\cite{ribeiro2016should}, Integrated Gradients (IG)~\cite{sundararajan2017axiomatic}, SmoothGrad (SG)~\cite{smilkov2017smoothgrad}, and Smooth Integrated Gradients (SIG). These methods are briefly described in Appendix \ref{appendix:xai-methods}. 
All explanation methods are implemented using standard and widely adopted libraries\footnote{\texttt{shap}: \url{https://github.com/shap/shap}; \texttt{lime}: \url{https://github.com/marcotcr/lime}; \texttt{captum}: \url{https://captum.ai/}, \texttt{pytorch}: \url{https://pytorch.org/}} with default hyperparameters to ensure a fair comparison.
SHAP explanations are computed using \texttt{TreeExplainer} for tree-based models and \texttt{KernelExplainer} with 100 background samples for other models, reporting attributions for the predicted class only. 
LIME explanations generate 5,000 perturbed samples, reporting the top 10 feature attributions. 
Integrated Gradients (IG) explanations are computed using 50 integration steps from a zero baseline, with gradients taken with respect to the predicted class logit. 
SmoothIG combines IG with input noise, averaging 50 noisy samples ($\sigma=0.1$) via \texttt{captum}'s \texttt{NoiseTunnel}. 
SmoothGrad averages gradients over 20 noisy input samples ($\sigma=0.1$), reduced to grayscale for image data.

\subsection{Perturbations}
\label{sec:perturbations}

To simulate realistic noise, out-of-distribution effects, and distribution shifts in the input space, we apply three types of natural perturbations with controlled intensity. First, \textit{gaussian noise}, which we apply to each feature independently, i.e.\
$\tilde{x}_i = x_i + \sigma \cdot \text{std}(X_i) \cdot \epsilon_i, \epsilon_i \sim \mathcal{N}(0, 1),$
where $\text{std}(X_i)$ is computed on the current input split (not training data) and 
$\sigma$ controls the perturbation strength.
Then, \textit{missing values}, where individual feature entries are randomly masked with probability $p$, followed by median imputation computed on the perturbed sample, simulating incomplete or corrupted measurements. Finally, \textit{permutation}, which randomly permutes a fraction $f$ of features across samples, breaking feature-target dependencies while preserving marginal distributions.

Additionally, for MLP models, we evaluate robustness under three types of gradient-based adversarial attacks: BIM~\cite{kurakin2018adversarial}, PGD~\cite{madry2017towards}, and C\&W~\cite{carlini2017towards}. These attacks represent worst-case adversarial distribution shifts that maximally disrupt predictions while remaining imperceptible to human observers. BIM is run for 10 iterations with strength $\epsilon$ under an $\ell_\infty$ constraint. The step size is set to $\alpha = 0.25\,\epsilon$, following $\alpha = \epsilon / n_{\text{iter}} \cdot 2.5$. 
PGD is run for 20 iterations with strength $\epsilon$ and $\ell_\infty$ constraint. The step size is set to $\alpha = 0.125\,\epsilon$ using the same scaling rule. Attacks are initialized with a random start, while only a single restart is used (no multi-restart search is performed).
C\&W uses an untargeted $\ell_2$ attack with fixed parameters $c$, $\kappa=0.0$, $n_{\text{iter}}=100$, and learning rate $0.01$.

\subsection{Evaluation Metrics}
\label{sec:stability}
To quantify explanation stability under input perturbations for tabular data, we use rank-based correlation measures, which are scale-invariant and emphasize relative feature importance. 
Given a clean input $\mathbf{x}$ and its perturbed counterpart  $\tilde{\mathbf{x}}$, we compute attributions $\boldsymbol{\phi}(\mathbf{x})$ and $\boldsymbol{\phi}(\tilde{\mathbf{x}})$, then rank features by absolute attribution magnitude: $R_i = \mathrm{rank}(|\phi_i(\mathbf{x})|)$ and $\tilde{R}_i = \mathrm{rank}(|\phi_i(\tilde{\mathbf{x}})|)$.
Specifically, we employ \textit{Spearman's $\rho$} for $d$ features:
\begin{align}
\rho(\boldsymbol{\phi}(\mathbf{x}), \boldsymbol{\phi}(\tilde{\mathbf{x}})) =
1 - \frac{6 \sum_{i=1}^{d} (R_i - \tilde{R}_i)^2}{d(d^2 - 1)}, \quad \in [-1, 1],
\label{eq:spearman}
\end{align}
and \textit{Kendall's $\tau$}:
\begin{align}
\tau(\boldsymbol{\phi}(\mathbf{x}), \boldsymbol{\phi}(\tilde{\mathbf{x}})) = 
\frac{C - D}{\binom{d}{2}}, \quad \in [-1, 1],
\label{eq:kendall-tau}
\end{align}
where $C = |\{(i,j): (R_i - R_j)(\tilde{R}_i - \tilde{R}_j) > 0\}|$ counts concordant pairs and $D = |\{(i,j): (R_i - R_j)(\tilde{R}_i - \tilde{R}_j) < 0\}|$ discordant.

While \textit{Spearman's $\rho$} captures global rank consistency, \textit{Kendall's $\tau$} is more sensitive to local rank inversions. We use Kendall's $\tau$ for the stability of feature rankings in individual explanations, and Spearman's $\rho$ for global trends where the overall monotonic change matters more than individual swaps.
For both metrics we use \texttt{scipy}'s implementation.

\medskip

In contrast to tabular data, images are inherently spatial and, therefore, we quantify stability using \textit{structural similarity (SSIM)}:
\begin{align}
\text{SSIM}(A, B) = 
\frac{(2\mu_A\mu_B + c_1)(2\sigma_{AB} + c_2)}
{(\mu_A^2 + \mu_B^2 + c_1)(\sigma_A^2 + \sigma_B^2 + c_2)}, \quad \in [0,1],
\label{eq:ssim}
\end{align}
where $\mu$, $\sigma^2$, $\sigma_{AB}$ denote local means (luminance), variances (contrast), and covariance (structure), and $c_1$, $c_2$ are stabilization constants.

\section{Experimental Results}
\label{sec:experiments}

This section reports experimental results evaluating the relationship between epistemic uncertainty and explanation stability, as well as the effectiveness of uncertainty-based gating.

\subsection{UQ Models performance}
\label{sec:experiment-performance}

Predictive performance and uncertainty statistics of all models are summarized in Table~\ref{tab:uq-summary}. Introducing uncertainty quantification does not materially affect predictive performance: F1 scores of UQ-enabled models closely match those of their deterministic counterparts across all datasets and architectures. 

\begin{table}[!b]
\centering
\caption{Predictive performance and uncertainty statistics. F1 (Non-UQ) corresponds to deterministic models without uncertainty estimation. Epistemic: mean test-set epistemic uncertainty; $\text{CV}_\text{epi}$: coefficient of variation measuring uncertainty dispersion. GBDTs epistemic is not evaluated (--).}
\label{tab:uq-summary}
\setlength{\tabcolsep}{3pt}
\fontsize{8pt}{10pt}\selectfont
\begin{tabular}{l l | c c | c c}
\hline
\textbf{Dataset} & \textbf{Model} & \textbf{F1 (Non-UQ)} & \textbf{F1 (UQ)} & \textbf{Epistemic} & \textbf{$\text{CV}_\text{epi}$} \\
\hline
\hline
Wine & LR        & 0.736 & 0.733 & 0.0004 & 1.40 \\
     & RF        & 0.810 & 0.810 & 0.083  & 0.50 \\
     & MLP       & 0.764 & 0.773 & 0.023  & 0.71 \\
     & LGBM      & 0.808 & 0.806 & --     & --   \\
     & CB        & 0.780 & 0.780 & --     & --   \\
\hline
Bean & LR        & 0.921 & 0.923 & 0.0001 & 3.71 \\
     & RF        & 0.925 & 0.925 & 0.0083 & 1.57 \\
     & MLP       & 0.936 & 0.935 & 0.026  & 1.41 \\
     & LGBM      & 0.930 & 0.930 & --     & --   \\
     & CB        & 0.931 & 0.931 & --     & --   \\
\hline
Rice & LR        & 0.921 & 0.923 & 0.0003 & 2.21 \\
     & RF        & 0.919 & 0.919 & 0.026  & 1.44 \\
     & MLP       & 0.930 & 0.921 & 0.0066 & 1.12 \\
     & LGBM      & 0.918 & 0.916 & --     & --   \\
     & CB        & 0.916 & 0.916 & --     & --   \\
\hline
Ecoli & LR        & 0.842 & 0.875 & 0.0016 & 2.25 \\
      & RF        & 0.845 & 0.845 & 0.0197 & 0.74 \\
      & MLP       & 0.869 & 0.869 & 0.044  & 0.71 \\
      & LGBM      & 0.824 & 0.809 & --     & --   \\
      & CB        & 0.841 & 0.841 & --     & --   \\
\hline
\end{tabular}
\end{table}

In addition to mean epistemic uncertainty, we characterize how uncertainty is distributed across samples using the \emph{epistemic coefficient of variation} 
$$\text{CV}_{\text{epi}} = \mathrm{std}(\mathcal{U}_{\text{epi}})~/~\mathrm{mean}(\mathcal{U}_{\text{epi}}).$$
Importantly, $\mathrm{CV}_{\mathrm{epi}}$ does not quantify how uncertain a model is on average, instead, it captures whether epistemic uncertainty is discriminative across the input space. High values indicate that uncertainty varies substantially between samples, separating regions where the model’s predictions are epistemically well-determined from regions where they are not. In contrast, low values indicate a near-uniform uncertainty landscape in which the model is similarly uncertain.
This distinction is crucial for uncertainty-aware explainability. Epistemic uncertainty can only serve as a validity signal for explanations if it can be used to meaningfully discriminate between samples. When epistemic uncertainty is highly scattered (high $\mathrm{CV}_{\mathrm{epi}}$), it localizes regions of epistemic indeterminacy, enabling stratification into reliable and unreliable explanation regimes. When dispersion is low (low $\mathrm{CV}_{\mathrm{epi}}$), epistemic uncertainty collapses to an almost uniform uncertainty landscape, where no such separation is possible.

The \textit{Dry Bean} and \textit{Rice} datasets exhibit higher epistemic variation, indicating heterogeneous model confidence and well-structured decision boundaries where epistemic uncertainty meaningfully separates samples. In contrast, lower epistemic variation in \textit{Wine Quality} and \textit{Ecoli} suggests that uncertainty is dominated by intrinsic data noise. In Wine, noise stems from subjective labeling, and in Ecoli from the small-sample, multi-class structure, creating a global uncertainty floor that limits the discriminative utility of epistemic uncertainty.

For LightGBM and CatBoost, native epistemic uncertainty is not reported due to architectural limitations; in subsequent experiments, epistemic uncertainty for these models is obtained via a Random Forest surrogate.

These observations provide context for the following experiments, which investigate when epistemic uncertainty can reliably predict explanation stability and when its utility is limited by dataset noise characteristics.

\subsection{XAI-UQ correlation analysis}
\label{sec:experiment-correlation}

In the following we examine whether epistemic uncertainty provides a reliable signal of explanation stability. Therefore, we analyze the relationship between changes in epistemic growth and degradation of post-hoc explanations under controlled perturbations. This analysis tests the central premise of uncertainty-aware explainability: that epistemic uncertainty increases precisely in regimes where explanations become unstable.
For each dataset, model, and XAI method, correlations are computed on a fixed test subset of size $n=\min(100,|X_{\text{test}}|)$ (sampled once with a fixed random seed) and shared across all perturbation strengths to ensure comparability.
The perturbation levels are denoted by $\lambda \in \Lambda$. The levels are set as follows for the different types:

\begin{itemize}[nosep,label=\textbullet]
    \item Gaussian noise scale $\Lambda \equiv \sigma \in \{0.01, 0.05, 0.1, 0.3, 0.5, 1.0, 2.0\}$,
    \item Missing values rate $\Lambda \equiv p \in \{0.01, 0.05, 0.1, 0.2, 0.3, 0.4, 0.5\}$,
    \item Feature permutation fraction $\Lambda \equiv f \in \{0.01, 0.02, 0.05, 0.1, 0.15, 0.2, 0.25\}$,
    \item BIM/PGD attack strength $\Lambda \equiv \epsilon \in \{0.01, 0.05, 0.1, 0.2\}$,
    \item C\&W regularization parameter $\Lambda \equiv c \in \{0.1, 1.0, 10.0\}$.
\end{itemize}

Explanation degradation (ED) is quantified by Kendall's $\tau$ (Eq.~\ref{eq:kendall-tau}) between attributions of clean inputs $\mathbf{x}_i$ and their perturbed counterparts $\tilde{\mathbf{x}}_i^{(\lambda)}$ at level $\lambda$, averaged across $n$ samples:

\begin{align}
XD(\lambda) = \frac{1}{n} \sum_{i=1}^{n} 
\tau\!\left(\boldsymbol{\phi}(\mathbf{x}_i), \boldsymbol{\phi}(\tilde{\mathbf{x}}_i^{(\lambda)})\right),
\end{align}

To capture the response of epistemic uncertainty to perturbations, we define epistemic growth (EG) as the relative increase\footnote{Alternatively, $EG(\lambda) = \frac{1}{n}\sum_{i=1}^{n} \mathcal{U}_{\text{epi}}(\tilde{\mathbf{x}}_i^{(\lambda)})$ -- absolute values yield identical XEC since Spearman's $\rho$ operates on ranks; the ratio aids human interpretability.} in mean epistemic uncertainty under perturbation:

\begin{align}
EG(\lambda) = \frac{\sum_{i=1}^{n} \mathcal{U}_{\text{epi}}(\tilde{\mathbf{x}}_i^{(\lambda)})}{\sum_{i=1}^{n} \mathcal{U}_{\text{epi}}(\mathbf{x}_i)},
\end{align}


Finally, for each perturbation type, we quantify the explanation-epistemic correlation (XEC) between XD and EG by computing the Spearman rank correlation $\rho$ (Eq.~\ref{eq:spearman}) across all perturbation levels $\Lambda$:

\begin{align}
XEC(\Lambda) = \rho\big(XD(\Lambda), EG(\Lambda)\big),
\label{eq:xai-uq-correlation}
\end{align}

A strongly negative correlation indicates that epistemic uncertainty consistently increases as explanations degrade. Throughout, we consider $XEC < -0.6$ as indicative of a strong negative association, suggesting that UQ reliably predicts XAI degradation.

\begin{figure}[!b]
    \centering
    \includegraphics[width=\linewidth]{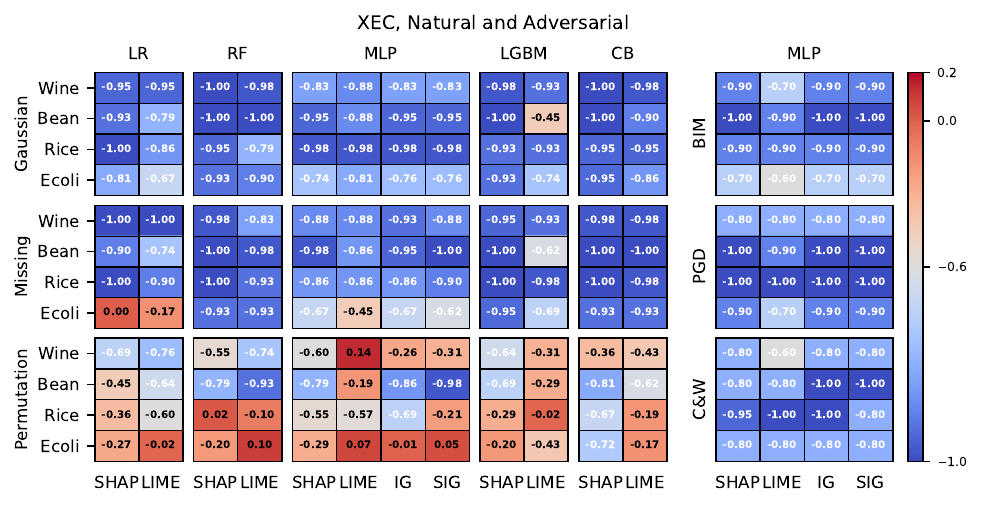}
    \caption{Correlation heatmaps of XAI stability and epistemic uncertainty (XEC) under natural (left) and adversarial (right) perturbations.
    Each cell reports Spearman correlation between epistemic growth (EG) and explanation degradation (XD) across perturbation strengths for a given dataset, model and XAI method.
    Blue and gray regions ($\text{XEC} < -0.6$) correspond to strong negative association, where UQ serves as a reliable proxy for XAI instability.
    }
    \label{fig:correlation-heatmaps}
\end{figure}

Figure~\ref{fig:correlation-heatmaps} summarizes the relationship between XAI stability and epistemic uncertainty across datasets, models, explanation methods, and perturbation regimes. Across nearly all settings, a strong negative association between EG and XD is observed, indicating that epistemic uncertainty systematically increases in regimes where explanations become fragile. This consistency across perturbation types and data modalities demonstrates that epistemic uncertainty captures a fundamental aspect of explanation robustness rather than a method- or perturbation-specific artifact.

Overall, SHAP yields stronger XEC than LIME, while Integrated Gradients consistently outperforms SmoothGrad among gradient-based methods. Across models and perturbation types, SHAP is the most reliable baseline, motivating its use in subsequent experiments.

Permutation perturbations yield the weakest correlations ($\text{XEC} > -0.6$ frequently). This behavior is expected, as feature permutation induces non-additive distribution shifts that explicitly disrupt feature-target dependencies.

For GBDTs (LightGBM and CatBoost), epistemic uncertainty was estimated via a RF surrogate. XEC remained strong and comparable to models with native epistemic estimates, indicating that the surrogate provides a sufficiently informative and stable epistemic signal.

Furthermore, dataset-specific effects are also observed. \textit{Ecoli} exhibits weaker correlations than Wine, Bean, and Rice, which is consistent with its small sample size, multi-class structure, and higher intrinsic noise.
These challenging characteristics limit the dispersion of epistemic uncertainty and, consequently, its ability to discriminate between stable and unstable explanation regimes.

Under adversarial perturbations, the overall XEC signal remains clearly negative. A mild weakening is observed for LIME, likely due to internal sampling noise partially reducing adversarial effects.

Together, these results establish epistemic uncertainty as a robust indicator for explanation instability across models, explanation methods, and perturbation methods. Based on these findings, subsequent experiments focus primarily on RF-based models, the Wine, Bean, and Rice datasets, SHAP explanations, and Gaussian noise as a canonical perturbation. Other models and perturbation types are retained selectively for comparative analysis.

\subsection{Stratified validation}
\label{sec:experiment-stratified}

In the following, we perform a stratified validation to assess whether epistemic uncertainty provides a meaningful proxy for explanation robustness at the level of individual predictions. 
While the previous correlation analysis established a global association between epistemic uncertainty and explanation instability (EG vs XD), we now test the stronger condition required for uncertainty-aware explainability: whether samples with higher epistemic uncertainty exhibit systematically stronger explanation degradation under input perturbations compared to low-epistemic samples.

For each sufficiently large dataset (Wine, Dry Bean, Rice), epistemic uncertainty is computed on the clean test set using the RF-based uncertainty model. The Ecoli dataset is excluded due to insufficient sample size for reliable stratification. Test samples are then stratified into three epistemic groups using quantile-based binning with equal-sized bins over the empirical epistemic uncertainty distribution, corresponding to low, medium, and high epistemic regimes. From each group, 50 samples are randomly selected.

For each selected sample, SHAP explanations are computed for the clean input and for the perturbed inputs obtained by adding Gaussian noise with $\sigma \in \{0.01, 0.05, 0.1\}$ across 10 noise seeds. Explanation stability is measured per sample using Kendall’s $\tau$ (Eq.~\ref{eq:kendall-tau}) between the rankings of absolute attributions and averaged across noise seeds.

\begin{figure}[!h]
    \centering
    \includegraphics[width=0.85\linewidth]{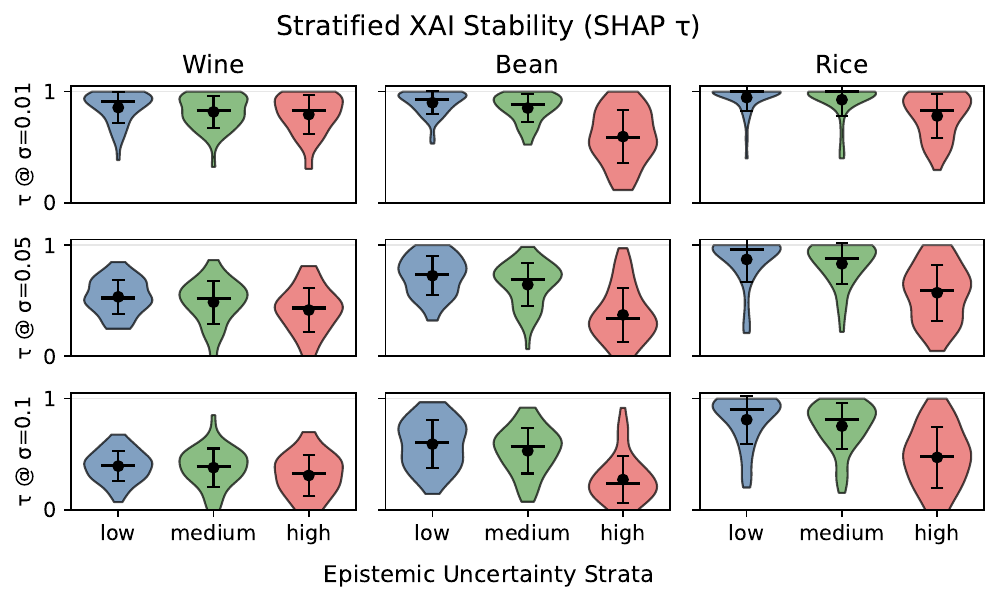}
    \caption{Stratified SHAP stability (Kendall’s $\tau$) across epistemic uncertainty bins under Gaussian noise ($\sigma$) for \texttt{low} (blue), \texttt{medium} (green), and \texttt{high} (red) epistemic strata. Violin plots show the distribution of stability values. Dots indicate means, horizontal bars the median, and vertical bars the standard deviation.}
    \label{fig:stratified-shap-stability}
\end{figure}

Figure~\ref{fig:stratified-shap-stability} reports stratified SHAP explanation stability across epistemic uncertainty strata (\texttt{low}, \texttt{medium}, \texttt{high}) under increasing Gaussian noise levels.
For each dataset and $\sigma$, violin plots show the distribution of SHAP $\tau$ values within each epistemic group.

Across all datasets and noise levels, we observe \texttt{low} $>$ \texttt{medium} $\gg$ \texttt{high} consistently.
As perturbation strength increases ($\sigma=0.01 \rightarrow 0.1$, top $\rightarrow$ bottom rows), overall $\tau$ decreases for all groups. Samples with \texttt{low} epistemic uncertainty exhibit the most stable explanations, \texttt{medium}-epistemic samples show moderate degradation, and \texttt{high}-epistemic samples consistently display substantially reduced stability. As perturbation strength increases ($\sigma = 0.01 \rightarrow 0.1$), explanation stability decreases across all groups, however, the $\tau$ degradation is markedly stronger for the \texttt{high} epistemic group.

Overall, these results provide sample-level evidence that epistemic uncertainty reliably serves as a proxy for explanation fragility: samples identified as epistemically uncertain are systematically associated with less stable SHAP explanations, justifying the use of epistemic uncertainty as a routing signal for explanation effort allocation.

\subsection{Recalling High-Quality Explanations}
\label{sec:experiment-gating}

In realistic deployment scenarios, the magnitude and nature of input perturbations are typically unknown. To simulate this uncertainty, we evaluate epistemic gating under  mixed perturbation conditions. Therefore, we consider 500 test samples over 10 noise levels $\sigma \in \{0.02\times k\}_{k=1}^{10}$. For each $\sigma$, we generate 5 perturbed versions per sample and average RF epistemic uncertainty (computed on perturbed inputs) and SHAP $\tau$ (Eq.~\ref{eq:kendall-tau}) across seeds, using clean SHAP ranks as the reference. The resulting mixed-noise population (5000 samples), formed by concatenating these per-$\sigma$ averaged samples, is used for evaluation.

\paragraph{Deferral-stability trade-off.}
Filtering by epistemic uncertainty introduces a fundamental trade-off: higher deferral rate $\nu$ (see Sec.~\ref{sec:cost}) yields higher-quality explanations but fewer of them.
A qualitative view of this trade-off, illustrating stable and unstable explanations as a function of epistemic uncertainty, is provided in Appendix~\ref{sec:epistemic-scatter}. To quantify this, we label samples as \textit{stable} if $\tau \geq 0.7$ and \textit{unstable} otherwise. We then evaluate epistemic-based filtering at various deferral rates $\nu \in [0.1, 0.9]$: at $\nu$, we reject samples with epistemic uncertainty above the $\nu$-th percentile of the epistemic distribution.

Table~\ref{tab:rejection-tradeoff} quantifies the precision-recall trade-off across deferral rates. In human-facing applications where explanation reliability is critical, precision is typically prioritized: at 50\% deferral, Bean and Rice achieve near-perfect precision (99.6\% and 100\%), while Wine reaches 73.5\%. Lower $\nu$ increases recall at the cost of precision, allowing practitioners to tune the trade-off based on their tolerance for unreliable explanations.

\begin{table}[!h]
\centering
\caption{Precision and recall for detecting stable explanations ($\tau \geq 0.7$) at different deferral rates $\nu$ under pooled noise conditions.}
\label{tab:rejection-tradeoff}
\setlength{\tabcolsep}{5pt}
\fontsize{8pt}{10pt}\selectfont
\begin{tabular}{l | c c | c c | c c}
\hline
 & \multicolumn{2}{c|}{\textbf{Wine}} & \multicolumn{2}{c|}{\textbf{Bean}} & \multicolumn{2}{c}{\textbf{Rice}} \\
\textbf{$\nu$} & \textbf{Prec.} & \textbf{Rec.} & \textbf{Prec.} & \textbf{Rec.} & \textbf{Prec.} & \textbf{Rec.} \\
\hline
\hline
90\% & 0.932 & 0.144 & 1.000 & 0.123 & 1.000 & 0.114 \\
70\% & 0.788 & 0.367 & 1.000 & 0.370 & 1.000 & 0.342 \\
50\% & 0.735 & 0.570 & 0.996 & 0.614 & 1.000 & 0.570 \\
30\% & 0.687 & 0.746 & 0.989 & 0.854 & 0.998 & 0.797 \\
10\% & 0.648 & 0.904 & 0.890 & 0.987 & 0.937 & 0.962 \\
\hline
\end{tabular}
\end{table}

\paragraph{Computational cost analysis.}
Beyond indicating explanation reliability, epistemic gating directly reduces computational cost by avoiding the computation of explanations for predictions that are epistemically ill-posed. The cost reduction depends on the relative cost of uncertainty estimation and explanation generation (see our dimensionality analysis in Section \ref{sec:cost}). Here we consider two representative scenarios:
\begin{itemize}
    \item[(1)] \textbf{RF with TreeSHAP}: Epistemic uncertainty (tree variance) is a byproduct of prediction, so $m/d \approx 0$ in Eq.~\eqref{eq:dim2}. Cost reduction simplifies to $q = 1/d + (1-\nu) \approx (1-\nu)$.
    \item[(2)] \textbf{MLP with LIME}: MC Dropout requires $m{=}50$ forward passes, while LIME requires $d{=}5000$ perturbations. From Eq.~\eqref{eq:dim1}, $m/d = 0.01$, making filtering highly cost-effective.
\end{itemize}

To quantify the trade-off between explanation stability and cost, we evaluate both configurations on the same mixed-noise population.
Table~\ref{tab:cost-benefit} reports mean stability and relative cost $q$ at varying deferral rates $\nu$. Filtering does not degrade explanation quality; mean stability increases as deferral rate grows, confirming that epistemic gating preferentially retains high-quality explanations.

For RF with TreeSHAP, rejecting $\nu=50\%$ of samples halves cost while improving stability from 0.740 to 0.777 (Wine), 0.821 to 0.937 (Bean), and 0.879 to 0.965 (Rice). Gains are largest for Bean and Rice where epistemic uncertainty is most discriminative (higher $\text{CV}_{\text{epi}}$, cf.\ Table~\ref{tab:uq-summary}).

For MLP with LIME, stability improvements are smaller; Wine exhibits near-constant stability across deferral rates, reflecting weaker epistemic signal. Nonetheless, $\nu=0.5$ achieves twofold cost reduction without degrading quality.

Overall, epistemic gating enables a controllable trade-off between deferral rate, quality, and computational cost. This also motivates Use Case 1: the same monotonic relationship supports routing high-uncertainty samples to more thorough XAI methods rather than applying expensive methods uniformly.

\begin{table}[!h]
\centering
\caption{Cost-benefit analysis for epistemic gating. Stability is reported as mean $\pm$ std of Kendall's $\tau$ for accepted samples. Relative cost $q$ follows Eqs.~\eqref{eq:dim1}--\eqref{eq:dim2}.}
\label{tab:cost-benefit}
\setlength{\tabcolsep}{5pt}
\fontsize{8pt}{10pt}\selectfont
\smallskip
\begin{tabular}{l c | c c | c c}
\hline
 & & \multicolumn{2}{c|}{\textbf{RF + TreeSHAP}} & \multicolumn{2}{c}{\textbf{MLP + LIME}} \\
\textbf{Dataset} & $\nu$ & \textbf{Stability} ($\tau$) & $q$ & \textbf{Stability} ($\tau$) & $q$ \\
\hline
\hline
Wine 
 & 70\%  & $0.797 \pm 0.121$ & 0.30 & $0.722 \pm 0.106$ & 0.31 \\
 & 50\%  & $0.777 \pm 0.130$ & 0.50 & $0.725 \pm 0.108$ & 0.51 \\
 & 30\%  & $0.758 \pm 0.141$ & 0.70 & $0.729 \pm 0.108$ & 0.71 \\
 & 0\%   & $0.740 \pm 0.149$ & 1.00 & $0.747 \pm 0.107$ & 1.00 \\
\hline
Bean 
 & 70\%  & $0.959 \pm 0.031$ & 0.30 & $0.712 \pm 0.120$ & 0.31 \\
 & 50\%  & $0.937 \pm 0.050$ & 0.50 & $0.692 \pm 0.144$ & 0.51 \\
 & 30\%  & $0.908 \pm 0.072$ & 0.70 & $0.669 \pm 0.156$ & 0.71 \\
 & 0\%   & $0.821 \pm 0.179$ & 1.00 & $0.635 \pm 0.165$ & 1.00 \\
\hline
Rice 
 & 70\%  & $0.976 \pm 0.034$ & 0.30 & $0.828 \pm 0.109$ & 0.31 \\
 & 50\%  & $0.965 \pm 0.042$ & 0.50 & $0.811 \pm 0.120$ & 0.51 \\
 & 30\%  & $0.947 \pm 0.058$ & 0.70 & $0.787 \pm 0.132$ & 0.71 \\
 & 0\%   & $0.879 \pm 0.150$ & 1.00 & $0.751 \pm 0.151$ & 1.00 \\
\hline
\end{tabular}
\end{table}

\subsection{Explanation Quality: Feature removal sensitivity}
\label{sec:experiment-quality}

The preceding experiments establish epistemic uncertainty as a reliable proxy for explanation stability. We now test via feature removal whether this extends to \emph{faithfulness}: low-epistemic explanations may not only be more stable but also more \emph{faithful}, identifying features that genuinely drive model predictions.
We address this explanation faithfulness through a feature removal experiment, by consecutively removing features that were correctly identified by SHAP as important features, which should substantially change the model's output. A complementary noise-attribution experiment, testing whether explanations focus on the signal rather than the noise, is reported in Appendix~\ref{sec:experiment-quality-noise}.

For each epistemic group (lowest, highest, random; 50 samples each), we compute SHAP attributions and successively remove the top-$k$ ($k \in \{1, \ldots, 5\}$) features by replacing their values with the training set median and measured the corresponding prediction shift with MSE.
%
To avoid saturation effects in the probability space, where highly confident predictions show negligible probability changes despite meaningful decision boundary shifts, we compute MSE in log-odds space rather than probability space.

\begin{figure}[!b]
    \centering
    \includegraphics[width=\linewidth]{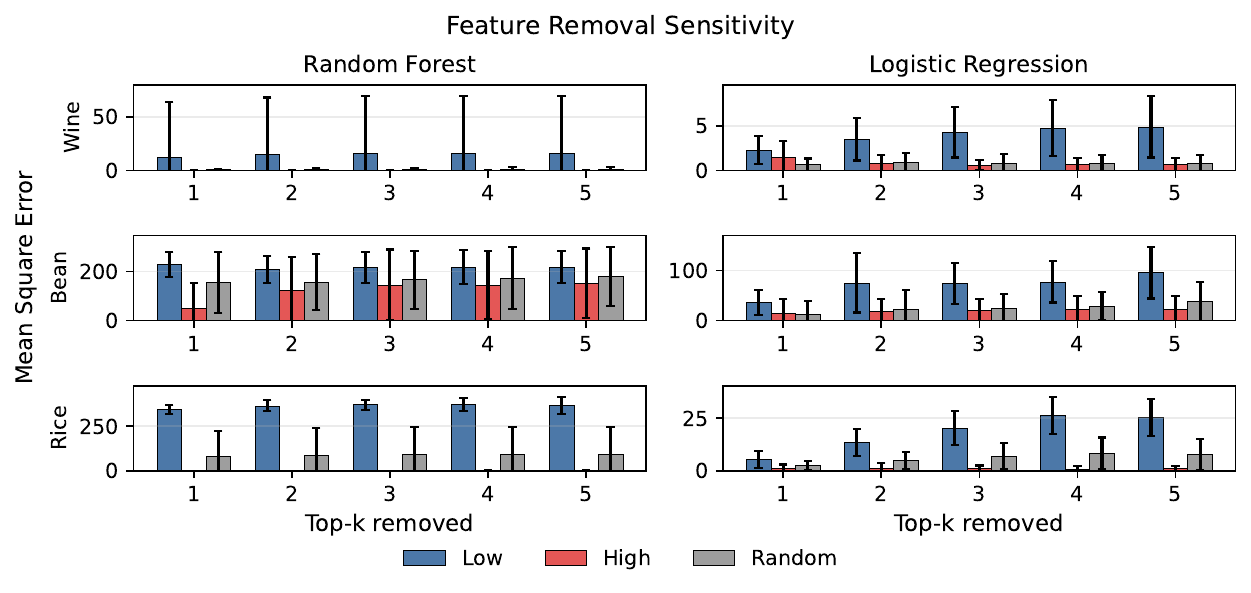}
    \caption{Prediction shift (MSE in log-odds space) after removing top-$k$ SHAP features by low- (blue), high- (red), and random-epistemic (gray) groups. Low-epistemic samples show larger shifts, indicating more faithful explanations.}
    \label{fig:feature-removal}
\end{figure}

From Figure~\ref{fig:feature-removal} we can observe that low-epistemic samples exhibit substantially higher prediction shifts than high-epistemic or random samples across all configurations. This indicates that SHAP attributions for low-epistemic samples correctly identify features that are genuinely important for model predictions. In contrast, high-epistemic samples show minimal prediction shifts, indicating that their explanations do not reliably capture the important features.

Together, the feature removal and noise attribution experiments demonstrate that epistemic uncertainty separates not only \emph{stable} from \emph{unstable} explanations, but also \emph{faithful} from \emph{unfaithful} ones. Low-epistemic explanations focus on signal features and reliably identify decision-relevant features, whereas high-epistemic explanations lack this semantic validity.

\subsection{Cross-Domain Validation: Image Classification}
\label{sec:experiment-images}

The preceding experiments establish a robust relationship between epistemic uncertainty and explanation stability for tabular data. To assess whether this relationship generalizes beyond tabular settings, we performed a cross-domain validation. We use the subset of PlantVillage dataset and CNN model described in Sections~\ref{sec:datasets} and \ref{sec:models}. Epistemic uncertainty is estimated via MC Dropout with 50 stochastic forward passes, computed as the variance of softmax probabilities. For explanations, we use Integrated Gradients and SmoothGrad (Sec.~\ref{sec:implementation}; SG with vanilla gradients). Attribution stability is measured using SSIM (Eq.~\ref{eq:ssim}) between clean and noisy saliency maps.
For the evaluation, we sample 100 test images and apply Gaussian noise at levels $\sigma \in \{0, 0.025, 0.05, 0.075, 0.1, 0.15, 0.2\}$. Similarly to XEC (Eq.~\ref{eq:xai-uq-correlation}, using SSIM as a measure of stability), for each $\sigma$, we compute mean epistemic uncertainty and mean SSIM, then measure their correlation across the different $\sigma$ using Spearman's $\rho$.

\begin{figure}[!h]
    \centering
    \includegraphics[width=0.85\linewidth]{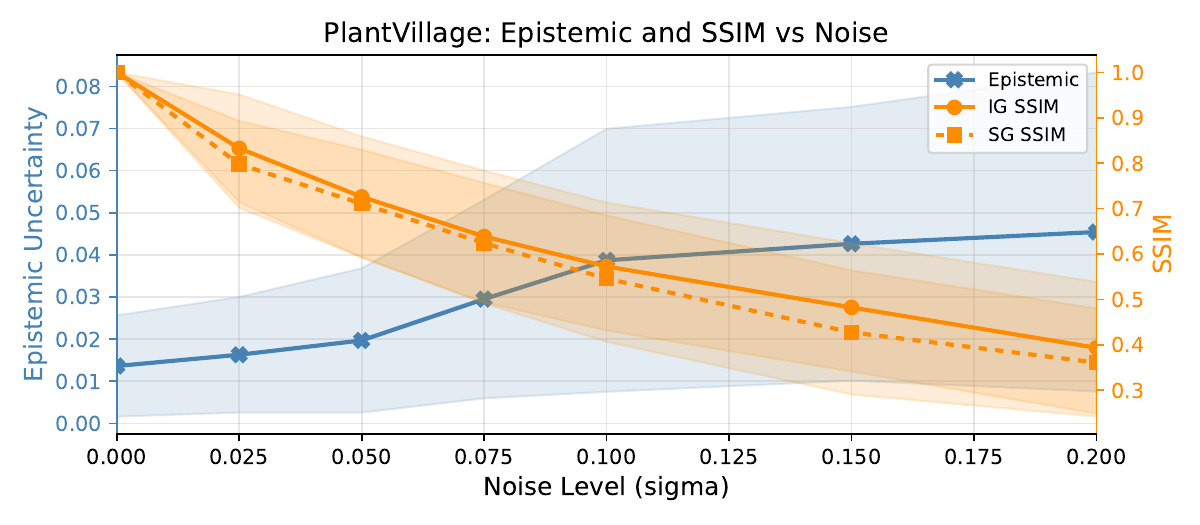}
    \caption{Epistemic uncertainty (left axis, blue) and SSIM stability (right axis, orange) means and standard deviations versus noise level (bottom axis). Axes are color-coded. As noise increases, uncertainty rises while SSIM decreases monotonically (XEC $\rho = -1.0$ for both IG and SG).}
    \label{fig:image-correlation}
\end{figure}

\paragraph{Aggregate correlation.}
Figure~\ref{fig:image-correlation} shows the relationship between epistemic uncertainty and attribution stability for increasing noise levels. Epistemic uncertainty increases monotonically with an increasing noise level, while SSIM decreases correspondingly for both explanation methods. The XEC (Eq.~\ref{eq:xai-uq-correlation}) between mean epistemic and mean SSIM across noise levels is $\rho = -1.0$ for both IG and SG, indicating perfect negative rank correlation.

\begin{figure}[!ht]
    \centering
    \includegraphics[width=\linewidth]{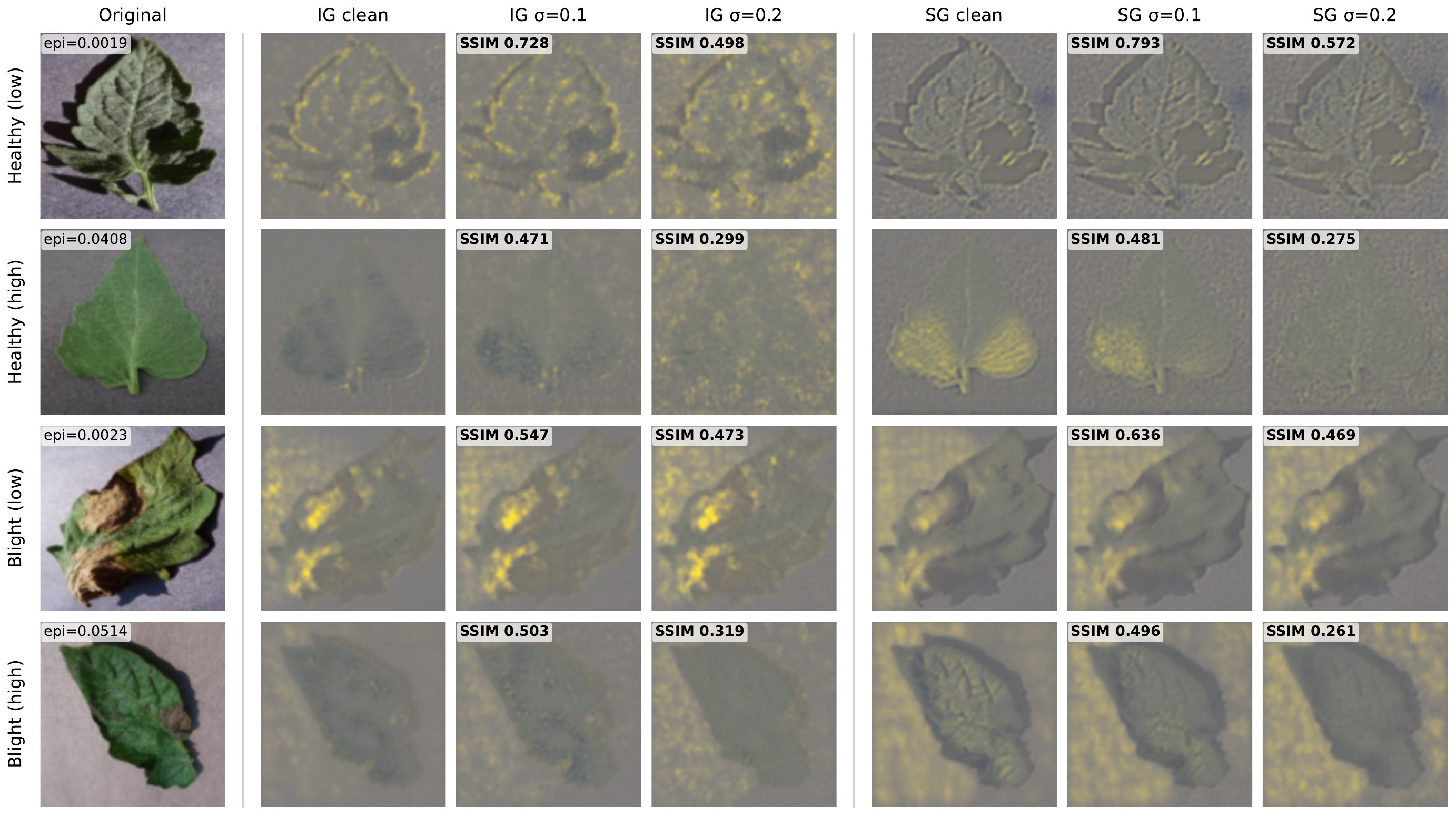}
    \caption{PlantVillage: IG and SG saliency maps for clean and noisy low-epistemic (rows 1, 3) and high-epistemic (rows 2, 4) samples. For visualization, heatmaps are lightly Gaussian‑smoothed to reduce pixel noise and emphasize structural patterns. Low-epistemic explanations maintain focused attribution on leaf structure and disease areas.}
    \label{fig:image-extremes}
\end{figure}

\paragraph{Qualitative analysis.}
Figure~\ref{fig:image-extremes} visualizes saliency maps at epistemic extremes. Samples with low epistemic uncertainty ($<2.5\times10^{-3}$) produce semantically coherent attributions (leaf venation for \enquote{healthy}, lesion regions for \enquote{blight}). This pattern is stable even under perturbation ($\text{SSIM}>0.47$ at $\sigma{=}0.2$). High-epistemic samples ($>4\times10^{-2}$) show diffuse saliency even on clean inputs and degrade rapidly ($\text{SSIM}\approx 0.3$ at $\sigma{=}0.2$), reinforcing uncertainty as a reliability indicator.

The consistency across data modalities, model architectures, explanation methods, and stability metrics confirms that epistemic uncertainty generalizes as a predictor of explanation quality.

\section{Discussion and Conclusion}
\label{sec:conclusion}

This work addresses a largely overlooked question in Explainable AI: not how to evaluate explanations once they have been produced, but how much computational effort explanation generation warrants. 
We introduce a framework that leverages epistemic uncertainty as a low-cost proxy for explanation reliability, enabling uncertainty-guided selective explanation generation across XAI methods. 
Experiments across four tabular datasets, five model architectures, multiple XAI methods, and diverse perturbation types demonstrate that epistemic uncertainty reliably predicts explanation degradation.
Stratified validation confirms that samples with high epistemic uncertainty exhibit systematically lower explanation stability, while quality experiments show that explanations of low-epistemic samples focus on signal features and faithfully capture model behavior. 
These findings generalize beyond tabular data as shown on an image classification task with CNNs and gradient-based explanations. 
Together these results establish epistemic uncertainty as a reliability signal for post-hoc explanation methods. Specifically, our framework supports two deployment modes:
First, epistemic uncertainty guides adaptive method selection, routing samples to low- or high-cost XAI methods based on expected reliability. Second, under constrained budget, it serves as a hard gate triggering explanation computation only when explanations are expected to be meaningful.

Our analysis also highlights the limitations of uncertainty-aware explainability. 
In datasets where epistemic uncertainty exhibits low dispersion, quantified by a low epistemic coefficient of variation, uncertainty only provides weak separation between stable and unstable XAI samples.
Permutation perturbations yield weaker XAI-UQ correlations due to non-additive distribution shifts that break feature dependencies. 
The uncertainty threshold is not universal across use cases and requires per-dataset calibration, though statistical approaches (e.g., percentile-based) offer reasonable defaults.
Finally, this work focuses exclusively on classification; extension to other tasks remains to be explored.

In practice, the value of epistemic gating depends on the context. For inexpensive methods like TreeSHAP, filtering may add unnecessary complexity, but for costly methods like LIME, or when combining multiple explainers, the savings become substantial. 
Lightweight RF epistemic surrogates work well when native epistemic estimation is unavailable or too expensive (e.g., GBDTs~\cite{malinin2021uncertainty}). However, the surrogate captures data-space uncertainty of a different model class rather than the target model's parameter uncertainty; in regions where RF and GBDT decision boundaries diverge, gating decisions may be unreliable.
The rejection rate is best treated as an engineering knob rather than a fixed threshold.
And even when full coverage is needed, uncertainty remains useful: it can accompany each explanation as a trust indicator, letting users judge reliability for themselves.

In summary, epistemic uncertainty separates explanations not only by their stability but also by their epistemic validity. Predictions with high epistemic uncertainty yield explanations that are fragile and unfaithful, whereas explanations based on low-epistemic uncertain predictions yield explanations that are both robust and faithful. This dual role, as both a stability predictor and a quality indicator, provides a principled foundation for uncertainty-aware explainability pipelines. We demonstrate the usefulness of our proposed framework in data modalities, model architectures, and explanation methods. It offers a practical tool for deploying reliable XAI either through adaptive method selection and epistemic gating in resource-constrained environments or as a continuous reliability indicator when full coverage is required.

\begin{credits}
\subsubsection{Acknowledgments}
This work was funded by the Federal Ministry of Research, Technology and Space through the project REFRAME (ref. 01IS24073B), which supports research on robustness, trustworthiness, and domain adaptation of foundation models, and the project DCropS4OneHealth (ref. 16LW0528K), which investigates causal links between diversified cropping systems, agrobiodiversity, food quality, and human health in large-scale on-farm experiments. Furthermore it was funded by SpinFert, one of the Soil Mission projects within the Horizon European program (ref. 101157265).

\subsubsection{Disclosure of Interests}
The authors have no competing interests to declare that are relevant to the content of this article.

\end{credits}
\bibliographystyle{splncs04}
\bibliography{ref}


\appendix

\section{Explanation Methods}
\label{appendix:xai-methods}

This appendix briefly describes the post-hoc model-agnostic attribution methods considered in this paper. Formulas for computing them are provided in Table~\ref{tab:xai-methods}.

\begin{table}[!h]
\centering
\caption{Equations of XAI methods used in this work.}
\label{tab:xai-methods}
\setlength{\tabcolsep}{5pt}
\fontsize{8pt}{10pt}\selectfont
\smallskip
\begin{tabular}{lll}
\toprule
\textbf{Method} & \textbf{Attribution} & \textbf{Approach} \\
\midrule
SHAP~\cite{lundberg2017unified} & 
$\phi_i = \sum_{S \subseteq \mathcal{F} \setminus \{i\}} \frac{|S|!(|\mathcal{F}|-|S|-1)!}{|\mathcal{F}|!}\big[f_{S \cup \{i\}} - f_S\big]$ & 
Shapley values \\[6pt]
LIME~\cite{ribeiro2016should} & 
$\phi_i = w_i^*$ from $\min_{\mathbf{w}} \sum_{\mathbf{z}} \pi_{\mathbf{x}}(\mathbf{z})(f(\mathbf{z}) - \mathbf{w}^\top \mathbf{z})^2$ & 
Local surrogate \\[6pt]
IG~\cite{sundararajan2017axiomatic} & 
$\phi_i = (x_i - x'_i) \int_{0}^{1} \frac{\partial f(\mathbf{x}' + \alpha(\mathbf{x} - \mathbf{x}'))}{\partial x_i} d\alpha$ & 
Path integration \\[6pt]
SG/SIG~\cite{smilkov2017smoothgrad} & 
$\phi_i = \frac{1}{N} \sum_{n=1}^{N} \phi_i^{\text{base}}(\mathbf{x} + \epsilon_n)$ & 
Noise averaging \\
\bottomrule
\end{tabular}
\end{table}

\textit{SHAP} computes Shapley values from cooperative game theory, where $\mathcal{F}$ denotes the feature set and $f_S(\mathbf{x})$ the model prediction with only features in $S$ present. TreeSHAP~\cite{lundberg2020trees} provides exact polynomial-time computation for tree models, while KernelSHAP~\cite{lundberg2017unified} approximates values for black-box models via weighted regression.
\textit{LIME}~\cite{ribeiro2016should} fits a weighted linear model to perturbed inputs, where $\pi_{\mathbf{x}}(\mathbf{z})$ weights samples by proximity. The learned coefficients $w_j^*$ serve as attributions.
\textit{Integrated Gradients (IG)}~\cite{sundararajan2017axiomatic} accumulates gradients along a straight path from baseline $\mathbf{x}'$ (typically zero) to   $\mathbf{x}$, satisfying completeness and sensitivity axioms.
\textit{SmoothGrad (SG)}~\cite{smilkov2017smoothgrad} and \textit{Smooth Integrated Gradients (SIG)} reduce noise by averaging attributions over $N$ noisy input copies, where $\epsilon_n \sim \mathcal{N}(0, \sigma^2 I)$ is Gaussian noise and $\phi^{\text{base}}$ is vanilla gradients (SG) or IG (SIG). We use SIG for tabular data and SG for images.

\section{Epistemic Uncertainty vs. XAI stability}
\label{sec:epistemic-scatter}

\begin{figure}[!h]
    \centering
    \includegraphics[width=0.95\linewidth]{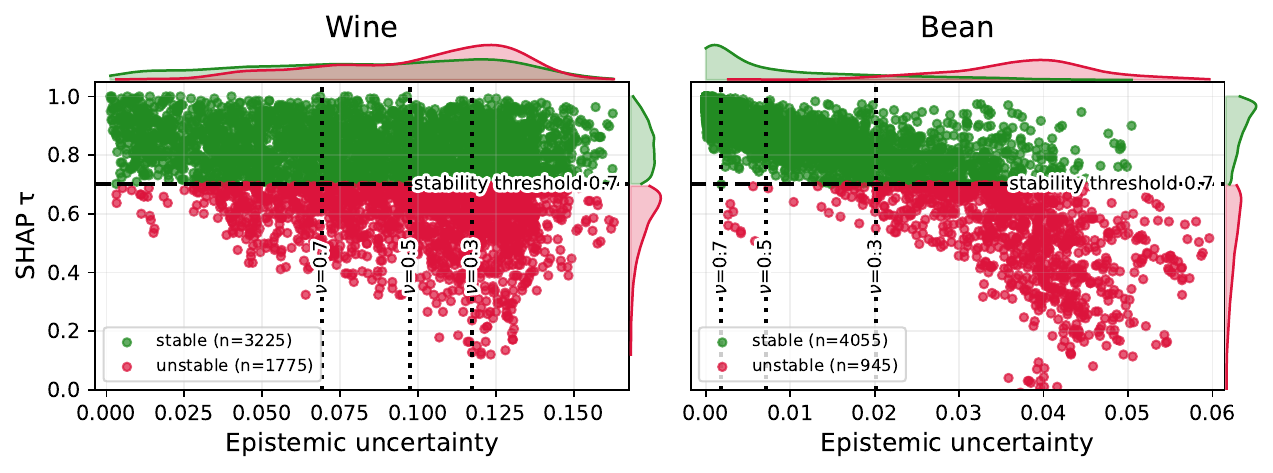}
    \caption{Epistemic uncertainty vs.\ SHAP stability (Kendall's $\tau$) under pooled noise conditions. 
    Green points denote stable explanations ($\tau \geq 0.7$), red points denote unstable. Vertical dashed lines show 30\%, 50\%, and 70\% deferral rates $\nu$. Marginal distributions illustrate class separation along each axis. 
    }.
    \label{fig:epistemic-scatter}
\end{figure}

This appendix provides a qualitative visualization of the coverage-stability trade-off discussed in Section~\ref{sec:experiment-gating}. In Figure~\ref{fig:epistemic-scatter} the relationship between epistemic uncertainty and explanation stability under pooled noise conditions for Wine and Bean is illustrated, representing weak and strong epistemic separation. Rice (not shown) behaves similarly to Bean. For the Bean dataset, we can observe a clear separation: stable samples concentrate at low epistemic values, and deferral rate $\nu=50\%$ retains almost exclusively stable explanations. Wine shows a more diffuse pattern due to higher intrinsic noise (lower $\text{CV}_{\text{epi}}$ cf.\ Table~\ref{tab:uq-summary}), though filtering still improves stability of the accepted set.

\section{Explanation Quality: Noise feature attribution}
\label{sec:experiment-quality-noise}

Here we complement the feature removal analysis (Section~\ref{sec:experiment-quality}) by testing whether epistemic uncertainty also affects the \emph{focus} of explanations.  Specifically, we examine whether high-epistemic samples spuriously attribute importance to irrelevant noise features. Therefore, each dataset is augmented with synthetic Gaussian noise features at varying ratios (1:1 to 10:1 noise-to-signal features for RF, 1:1 to 3:1 for LR). Models are retrained on the augmented data, and SHAP explanations are computed for three stratified epistemic groups: low, high, and random (50 test samples per group).

Attribution focus is quantified by the \emph{signal mass} -- the fraction of total absolute SHAP attribution assigned to original (signal) features:
\begin{align}
    \text{Signal Mass} = \frac{\sum_{i \in \mathcal{S}} |\phi_i|}{\sum_{i=1}^{d} |\phi_i|},
\end{align}
where $\mathcal{S}$ denotes the set of signal feature indices. Higher signal mass indicates that explanations correctly focus on predictive features rather than noise.

\begin{figure}[!h]
    \centering
    \includegraphics[width=0.9\linewidth]{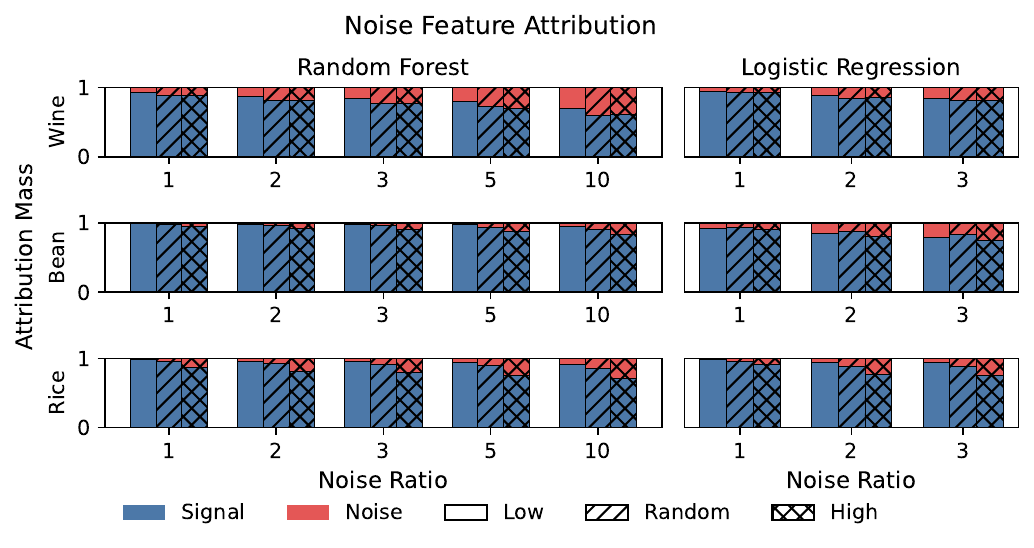}
    \caption{Attribution mass on signal (blue) vs. noise (red) features across epistemic groups (solid: low, hatched: random, cross-hatched: high) and noise feature ratios. Low-epistemic explanations maintain focus on signal features better.}
    \label{fig:noise-attribution}
\end{figure}

Figure~\ref{fig:noise-attribution} reports signal and noise attribution mass across epistemic groups, noise ratios, datasets, and models. Low-epistemic samples consistently maintain higher signal mass, indicating that their explanations better focus on predictive features. In contrast, high-epistemic samples show substantial drift toward noise features. Random samples fall between the two groups, as expected.
These results explain \emph{why} low-epistemic explanations are more stable: they rely on genuine data patterns rather than spurious correlations with noise. When perturbations are applied, explanations grounded in signal features remain consistent, while those drifting toward noise are inherently fragile.

\end{document}